\documentclass[a4paper,fleqn]{cas-dc}

\usepackage{algorithm}
\usepackage{algorithmic}
\usepackage[round]{natbib}
\usepackage{subcaption}
\usepackage{pifont}
\usepackage{bm}
\usepackage{comment}

\definecolor{grey}{RGB}{205,205,205}

\def\tsc#1{\csdef{#1}{\textsc{\lowercase{#1}}\xspace}}
\tsc{WGM}
\tsc{QE}

\begin{document}
\let\WriteBookmarks\relax
\def\floatpagepagefraction{1}
\def\textpagefraction{.001}
\captionsetup[figure]{labelfont={bf},labelformat={default},labelsep=period,name={Fig.}}

\shorttitle{A remote sensing benchmark for multi-modal semantic segmentation}    

\shortauthors{Zhitong Xiong, et al.}  

\title [mode = title]{GAMUS: A Geometry-aware Multi-modal Semantic Segmentation Benchmark for Remote Sensing Data}



%


\author[1]{Zhitong Xiong}[type=editor,
			          auid=1,
			          bioid=1,
			          prefix=,]
\ead{zhitong.xiong@tum.de}
\cormark[1]

\author[1,2]{Sining Chen}[type=editor,
					  auid=2,
					  bioid=2,
					  prefix=,]
\ead{sining.chen@tum.de}

\author[1,2]{Yi Wang}[type=editor,
					  auid=4,
					  bioid=4,
					  prefix=,]
\ead{yi4.wang@tum.de}

\author[1]{Lichao Mou}[type=editor,
					  auid=4,
					  bioid=4,
					  prefix=,]
\ead{lichao.mou@tum.de}

\author[1]{Xiao Xiang Zhu}[type=editor,
					  auid=5,
					  bioid=5,
					  prefix=,]
\ead{xiaoxiang.zhu@tum.de}
\cormark[1]






\affiliation[1]{organization={Chair of Data Science in Earth Observation, Technical University of Munich},
				city={Munich},
				postcode={80333},
				country={Germany}}
				
\affiliation[2]{organization={Remote Sensing Technology Institute, German Aerospace Center},
				city={Weßling},
				postcode={82234},
				&country={Germany}}

				









\cortext[1]{Corresponding author}



\begin{abstract}
Geometric information in the normalized digital surface models (nDSM) is highly correlated with the semantic class of the land cover. Exploiting two modalities (RGB and nDSM (height)) jointly has great potential to improve the segmentation performance. However, it is still an under-explored field in remote sensing due to the following challenges. First, the scales of existing datasets are relatively small and the diversity of existing datasets is limited, which restricts the ability of validation. Second, there is a lack of unified benchmarks for performance assessment, which leads to difficulties in comparing the effectiveness of different models. Last, sophisticated multi-modal semantic segmentation methods have not been deeply explored for remote sensing data. To cope with these challenges, in this paper, we introduce a new remote-sensing benchmark dataset for multi-modal semantic segmentation based on RGB-Height (RGB-H) data. Towards a fair and comprehensive analysis of existing methods, the proposed benchmark consists of 1) a large-scale dataset including co-registered RGB and nDSM pairs and pixel-wise semantic labels; 2) a comprehensive evaluation and analysis of existing multi-modal fusion strategies for both convolutional and Transformer-based networks on remote sensing data. Furthermore, we propose a novel and effective Transformer-based intermediary multi-modal fusion (TIMF) module to improve the semantic segmentation performance through adaptive token-level multi-modal fusion.
The designed benchmark can foster future research on developing new methods for multi-modal learning on remote sensing data. Extensive analyses of those methods are conducted and valuable insights are provided through the experimental results. Code for the benchmark and baselines can be accessed at \url{https://github.com/EarthNets/RSI-MMSegmentation}.
\end{abstract}
 



\maketitle

\section{Introduction}
	Semantic segmentation, aiming at assigning semantic labels for each image pixel, is a fundamental and long-standing goal of both the computer vision (CV) and remote sensing (RS) fields \citep{zhu2017deep,kuznietsov2017semi,mou2018im2height,xiong2022earthnets}. With the advent of deep learning techniques, RGB image-based semantic segmentation has attracted great research attention, and significant progress has been made on performance \citep{long2015fully, xie2021segformer, liu2021swin}. Despite the richness of texture information in RGB images, semantic segmentation models often face challenges in extracting discriminative features from them, owing to the inherent limitations of 2D RGB representations. Considering this problem, multi-modal image segmentation is becoming more and more popular in both the CV and RS communities. Multi-modal data such as RGB-Depth (RGB-D) \citep{xiong2021ask} and RGB-Thermal (RGB-T) \citep{li2019rgb}, contain richer information compared with traditional RGB images, and are widely used for many tasks \citep{DBLP:conf/eccv/WangN18,DBLP:conf/cvpr/ChengCLZH17, jiang2018multimodal, xiong2020msn}. Multi-modal data, which incorporates additional modalities beyond RGB, has shown promising results in achieving significantly improved performance compared to using RGB data alone.
	Remote sensing (RS) data presents several rich data modalities beyond RGB images, such as hyperspectral (HS), multi-spectral (MS), light detection and ranging (LiDAR), normalized digital surface model (nDSM), and synthetic aperture radar (SAR). As a result, multi-modal or multi-sensor data fusion has become a vital area of research in the remote sensing community.
 
		\begin{figure*}
		\centering
		\includegraphics[width=0.984\linewidth]{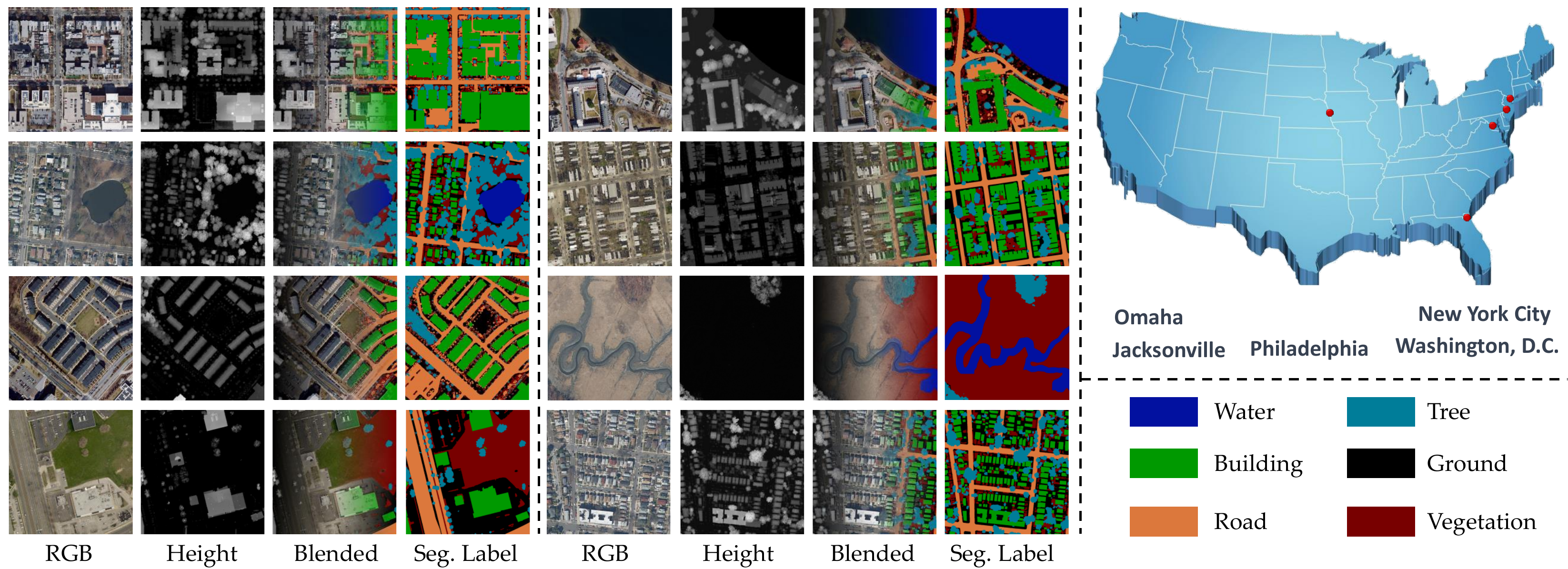}
		\caption{Example images of the GAMUS dataset. Images from left to right are the RGB modality, the nDSM modality, the blending visualization image, and the segmentation label.}
		\label{fig:globe}
	\end{figure*}
	There are various modalities widely used in RS for different Earth observation applications. The development of multi-modal benchmark datasets is critical for advancing research in multi-modal learning methods. However, for semantic segmentation on remote sensing (RS) data, existing multi-modal benchmark datasets have several limitations, given the application areas and data acquisition cost. The foremost limitation is the limited spatial resolutions of the available multi-modal datasets. While Sentinel-1 and Sentinel-2 images are widely-used satellite data with multiple modalities, their low spatial resolution fails to provide detailed land-use and land-cover information. The second limitation pertains to the limited geospatial coverage of existing multi-modal datasets. Although the ISPRS Potsdam and Vaihingen datasets have a higher spatial resolution of 5 cm (0.05m), obtaining such high-resolution images for large-scale real-world applications could be cost-prohibitive. The third limitation of the current multi-modal benchmark datasets is the lack of a unified benchmark platform to enable comprehensive and fair comparison of different multi-modal learning methods.

    Similar to the RGB-D segmentation task, several works \citep{zheng2021gather,audebert2018beyond,liu2019semantic} have proven that using the geometric information in nDSM leads to a higher segmentation performance on remote sensing data. However, as presented in Table \ref{T1}, existing multi-modal datasets that contain RGB and nDSM pairs are relatively small. This makes it difficult to faithfully compare the effectiveness of different types of multi-modal representation learning methods. These limitations motivate us to build a new multi-modal semantic segmentation dataset to enable a fair and unified evaluation of different multi-modal segmentation methods. The proposed Geometry-Aware Multi-modal Segmentation (GAMUS) dataset contains images with a resolution of 0.33m, which is higher enough to be used in many real-world applications. The GAMUS dataset contains two data modalities: RGB images and normalized digital surface models (nDSM). Since nDSM indicates the height of ground objects, we use the height modality to represent the nDSM data.   
    nDSM has been broadly provided by many cities owing to its importance in 3D city modeling. Unlike depth images in CV datasets, different types of land covers usually have unique height attributes. Thus, the geometric information contained in height maps is highly correlated to the semantic information \citep{kunwar2019u, mahmud2020boundary}. In other words, compared with other modalities, height information from nDSM has great potential in improving the segmentation performance of high-resolution remote sensing images. 
	\begin{table*}
		\caption{Statistics comparison with existing RGB-H datasets. Note that the test set of DFC 19 is not publicly accessible.}
		\label{T1}
		\scalebox{0.85}{
			\begin{tabular}{c|c|c|c|c|c|c|c}
				\hline \hline
				Datasets              & \#Total Tiles & Tile Size            & \#Cities          & \#Training          & \#Validation        & \#Test              & \#Class Labels \\ \hline
				Potsdam \citep{rottensteiner2014isprs}       & 38            & 6000 × 6000          & Single            & 24 tiles            & 0                   & 14 tiles            & 6              \\
				Vahingen \citep{rottensteiner2014isprs}      & 33            & 2500 × 2000          & Single            & 16 tiles            & 0                   & 17 tiles            & 5              \\
				DFC 19 \citep{c6tm-vw12-19}      & 2783            & 1024 × 1024          & Multiple            & 2783 tiles            & 100 tiles                  & ---            & 6              \\
				GeoNRW \citep{bosch2019semantic}      & 33            & 2500 × 2000          & Multiple            & 16 tiles            & 0                   & 17 tiles            & 5              \\
				Zeebruges \citep{yokoya2018open}      & 9             & 10,000 × 10,000        & Single            & 5 tiles             & 0                   & 2 tiles             & \textbf{8}     \\
				Augsburg \citep{hong2021multimodal}   & 1             & 332 × 485            & Single            & 761 pixels          & 0                   & 77,533 pixels        & 7              \\ \hline
				\textbf{GAMUS (Ours)} & \textbf{11,507} & {1024 × 1024} & \textbf{Multiple} & \textbf{6304 tiles} & \textbf{1059 tiles} & \textbf{4144 tiles} & 6              \\ \hline \hline
		\end{tabular}}
	\end{table*}

	Although extensive works have been proposed to make better use of the diverse information contained in different modalities \citep{hong2021multimodal}, there is still a lack of comprehensive benchmarking of existing multi-modal learning methods on RS data. There are many widely used strategies for multi-modal data fusion, including early fusion, feature-level fusion, late-fusion, and Transformer-based token fusion. However, it is still not clear which one is more suitable for the pixel-level semantic segmentation task on RGB-H (Height) data. Thus, the great potential of the nDSM modality is heavily overlooked by existing works.
    
    \begin{figure*}
		\centering
		\includegraphics[width=0.92\linewidth]{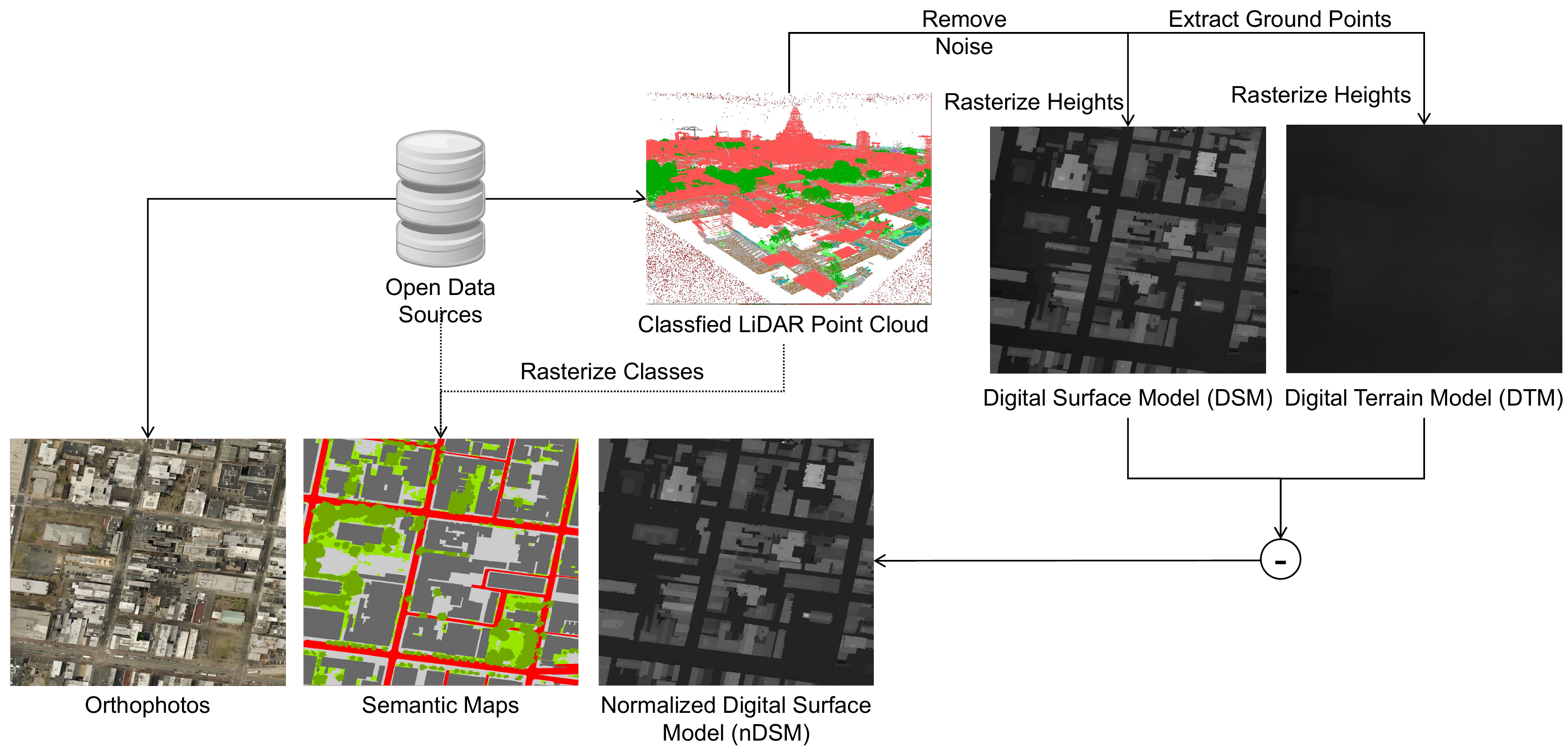}
		\caption{Data collection and processing of the GAMUS dataset.}
		\label{fig2}
    \end{figure*}

    Considering this problem, in this work, we propose a simple yet effective Transformer-based Intermediary Multi-modal Fusion (TIMF) Module. TIMF utilizes an intermediary learnable token to fuse the features of RGB and height modalities via the self-attention mechanism. To provide comprehensive benchmarking results, we conduct extensive experiments to evaluate Convolutional Neural Networks (CNN) and Transformer-based methods and different variants of fusion strategies. We believe that the benchmark dataset and released codebase can foster future research on evaluating and developing new multi-modal learning methods for Earth observation applications. We summarize the main contributions of this paper below.
	\begin{enumerate}
		\item A large-scale dataset (GAMUS) containing co-registered RGB and nDSM pairs and pixel-level semantic labels is introduced, which contains data from five different cities;
		\item Both CNN and Transformer-based multi-modal learning models with different modal-fusion strategies are compared and analyzed on remote sensing data. which enables a fair and comprehensive performance comparison and evaluation;
		\item A novel Transformer-based Intermediary Multi-modal Fusion (TIMF) module is proposed for the adaptive fusion of RGB and Height data, which achieves state-of-the-art segmentation performance. 
        \item The proposed datasets and benchmarking results can provide useful insights and spark novel ideas for developing new multi-modal segmentation methods for RS data.
	\end{enumerate}

 The remainder of this paper is structured as follows. Section \ref{RelatedWork} provides an overview of the related methods of multi-modal learning and existing multi-modal datasets for remote sensing (RS) data. Section \ref{GAMUS} presents detailed information about the proposed GAMUS dataset. In Section \ref{Methods}, we describe the CNN and Transformer-based multi-modal learning methods with different fusion strategies, along with the designed Transformer-based Intermediary Multi-modal Fusion (TIMF) module. Finally, in Section \ref{Experiments}, we present and analyze the benchmarking results obtained through comprehensive and fair comparisons of the proposed models on the GAMUS dataset.
 
\section{Related Work}
\label{RelatedWork}
	\paragraph{Multi-modal Representation Learning for Computer Vision} 
	Multi-modal feature learning is commonly studied in RGB-D image-based scene recognition \citep{yuan2019acm}, semantic segmentation \citep{ShapeConv,MFNet,SA-Gate,SGNet}, object detection \citep{gupta2014learning,fan2020rethinking}, and action recognition \citep{zhang2016rgb}. Multi-modal fusion is the key to designing multi-branch deep networks. Existing methods usually conduct fusion at the image level (early fusion) or the feature level (intermediate fusion) \citep{DBLP:journals/corr/abs-1806-01054}. FuseNet \citep{Hazirbas2016FuseNet} and RedNet \citep{DBLP:journals/corr/abs-1806-01054} summed RGB and depth features to obtain multi-modal representations for RGB-D semantic segmentation. Multi-level feature fusion \citep{Park_2017_ICCV} was designed to extend the residual learning idea of RefineNet \citep{Lin:2017:RefineNet} for RGB-D image segmentation. Similarly, ACNet \citep{DBLP:journals/corr/abs-1905-10089} and Gated Fusion Net \citep{DBLP:conf/cvpr/ChengCLZH17} were proposed to adaptively fuse features of different modalities for image segmentation. PSTNet \citep{PSTNet} and RTFNet \citep{RTFNet} were proposed to utilize long wave infrared (LWIR) imagery as a viable supporting modality for semantic segmentation using multi-modal learning networks. 
	\begin{figure*}
		\centering
		\includegraphics[width=0.99\linewidth]{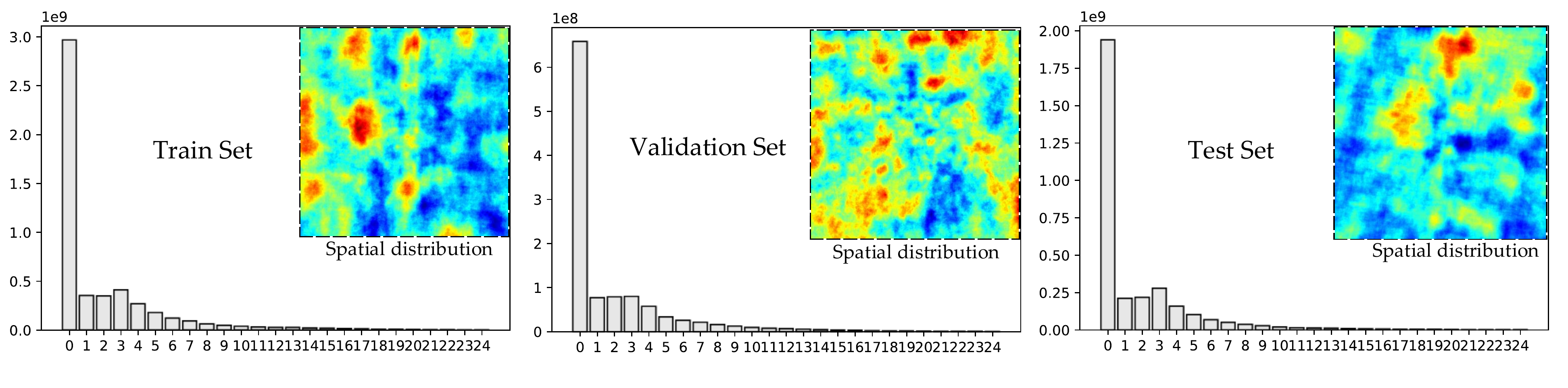}
		\caption{Statistics of height values of the GAMUS dataset. Both the histograms and spatial distributions are presented. A long-tailed distribution can be clearly observed.}
		\label{Hstat}
	\end{figure*}
	\paragraph{Multi-modal Representation Learning for Remote Sensing} 
	For remote sensing data, \citep{kampffmeyer2016semantic} designed a multi-modal deep network with an early deep fusion architecture by stacking all modalities as the input. A promising performance has been achieved on the semantic segmentation of urban images. Audebert et al. \citep{audebert2017joint} proposed to combine the optical and OpenStreetMap data using a two-stream multi-modal learning network to improve the segmentation performance. Exploring the combination of Multi-spectral images (MSI) and Lidar is closer to our work. Audebert et al. \citep{audebert2018beyond} introduced a SegNet-based multi-modal fusion architecture for the segmentation of urban scenes. Cross-modality learning (CML) was investigated by \citep{hong2020x,hong2020learning}. In \citep{hong2020learning}, a cross-fusion strategy was proposed for learning multi-modal features with more balanced weight contributions from different modalities. Liu et al. \citep{liu2019semantic} designed a high-order conditional random field (CRF) based method for the fusion of optical and Lidar predictions in a late-fusion manner. G2GNet \citep{zheng2021gather} was proposed to combine the complementary merits of RGB and auxiliary modality data using the introduced gather-to-guide module.
	
	\paragraph{Remote Sensing Datasets for Multi-modal Semantic Segmentation}
	The two widely used high-resolution semantic segmentation benchmark datasets in remote sensing are the ISPRS Potsdam and Vaihingen datasets \citep{rottensteiner2014isprs}. As compared in Table \ref{T1}, the  Potsdam  dataset  contains 38 tiles with about 6,000$\times$6,000 image resolution. Usually, 24 tiles are used for training and the rest 14 tiles for testing. The Vaihingen dataset contains 33 tiles with an image resolution of about 2,500$\times$2,000. These tiles are officially split into two subsets, of which 16 tiles are used for training and 17 tiles for testing.
    The DFC19 dataset is a remote sensing dataset that contains 2783 images of 1024X1024 resolution. It is designed for the Deep Learning for Semantic Segmentation of Urban Scenes challenge, which aims to advance the state-of-the-art in semantic segmentation of urban scenes using multi-spectral and LiDAR data.
    The GeoNRW dataset is a remote sensing dataset that contains 7783 images with pixel-level annotation. It includes 10 different land cover classes. The dataset includes RGB and nDSM data.
    The Zeebruges dataset \citep{yokoya2018open} was acquired using an airborne platform flying over the urban and harbor areas of Zeebruges, Belgium. The dataset contains seven separate tiles and each tile is with a 10,000 × 10,000 image resolution. Five tiles are used for training, and the remaining two tiles are for testing. Hong et al. \citep{hong2021multimodal} introduced a multi-modal dataset with Hyperspectral (HS), SAR, and nDSMs for the classification of remote sensing data. 
 
    \begin{figure}
		\centering
		\includegraphics[width=0.95\linewidth]{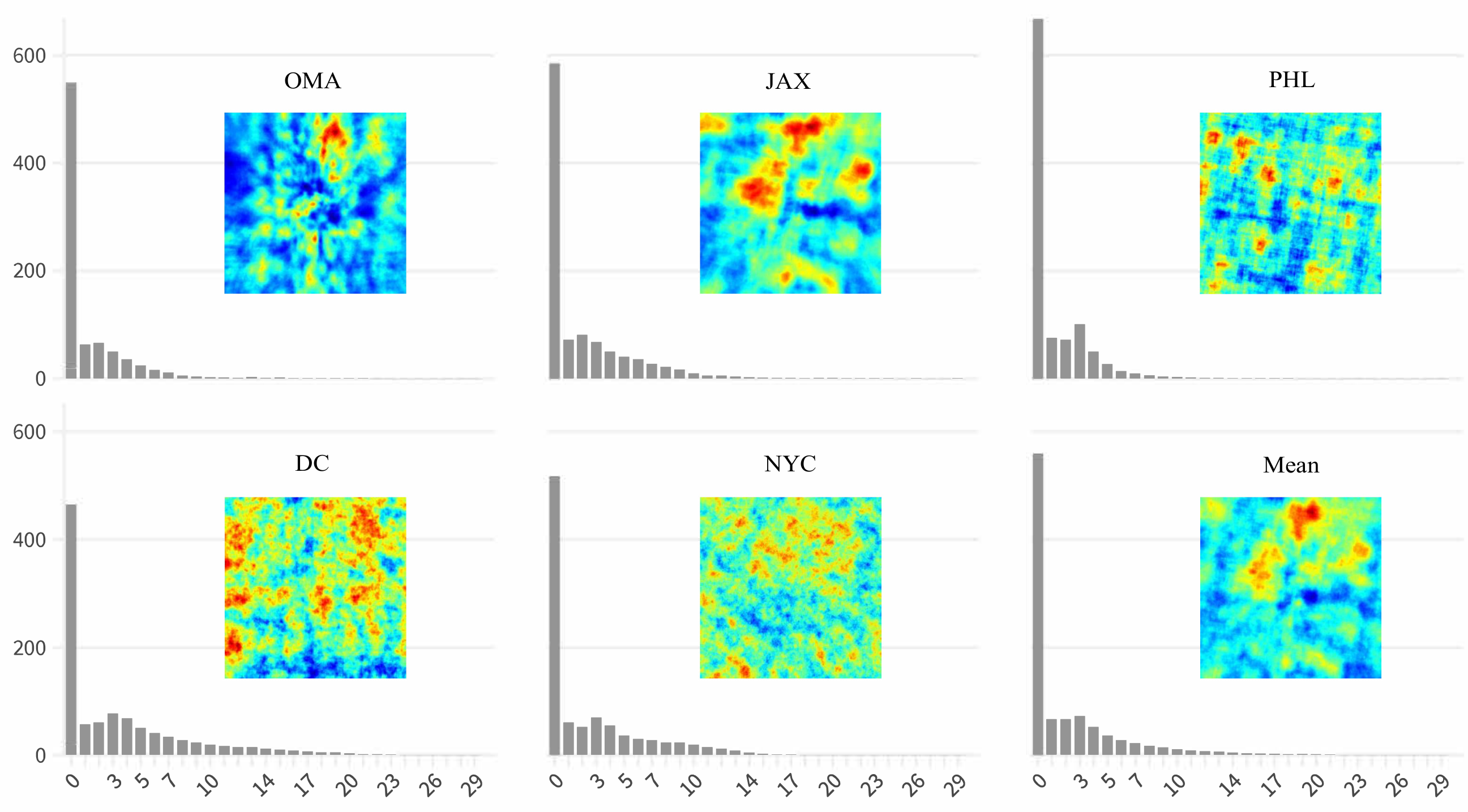}
		\caption{Statistics of height values of the GAMUS dataset. Both the histograms and spatial distributions are presented. A long-tailed distribution can be clearly observed.}
		\label{HCity}
    \end{figure}
	The US3D \citep{bosch2019semantic} dataset is collected from public data. Although it is large in volume (more than 700 GB) and contains data for many different tasks, it is not designed dedicated to multi-modal segmentation tasks. In contrast, the GAMUS dataset is smaller in volume, easier to use, and covers more cities. We provide a standard dataloader, detailed instructions, strong baselines, and extensive benchmark results for multi-modal learning. The DFC 19 \citep{c6tm-vw12-19} dataset is derived from US3D, which is smaller in scale.

    \begin{figure}
		\centering
		\includegraphics[width=0.92\linewidth]{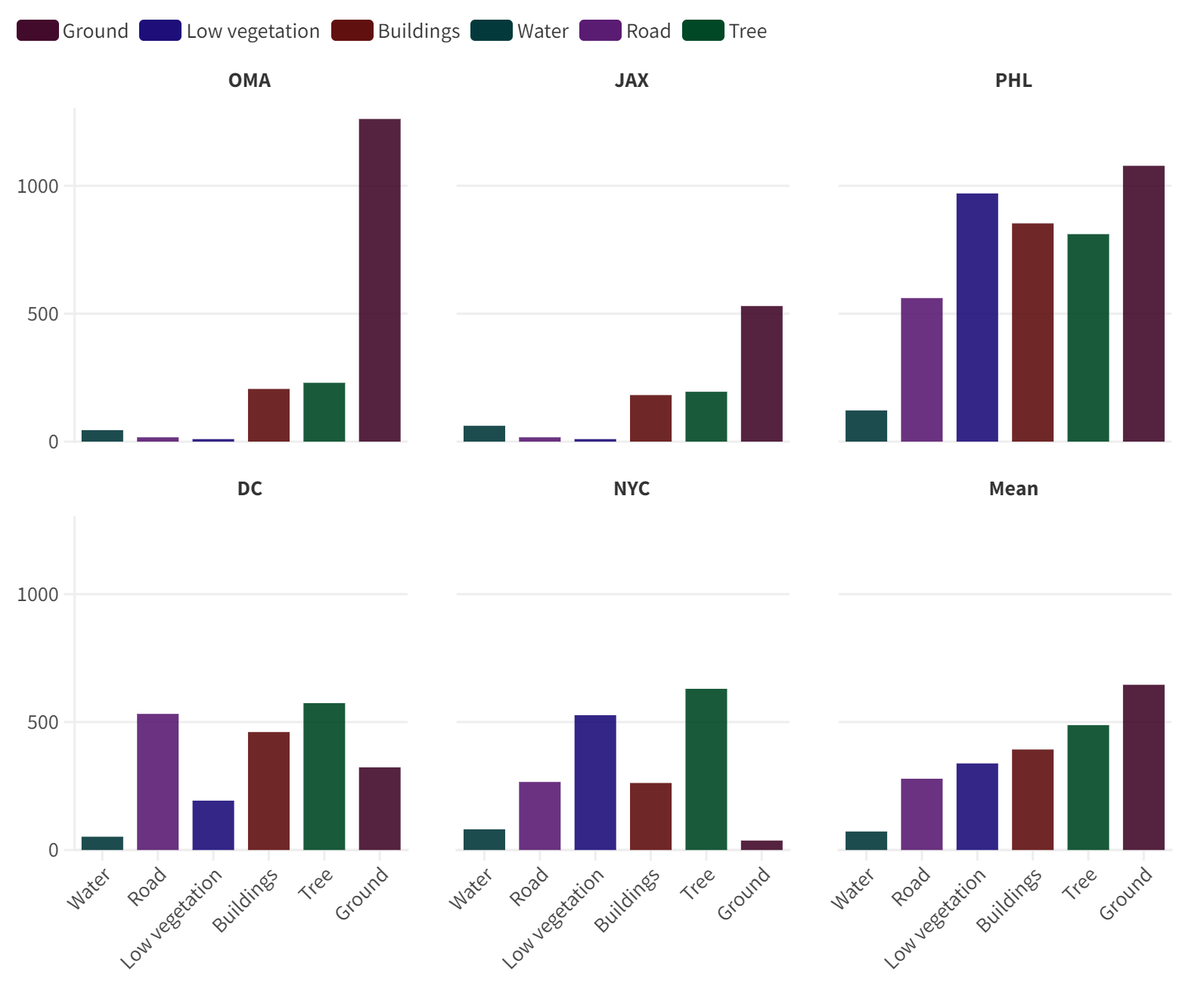}
		\caption{Statistics of the semantic class of the GAMUS dataset. }
		\label{HCLS}
    \end{figure}

\section{The GAMUS Dataset}
    \label{GAMUS}
	\subsection{Data collection}
	High-resolution orthophotos, semantic maps, and nDSM (height) are derived and processed from open Data DC catalog (https://opendata.dc.gov) and open data in the Philadelphia region (https://www.opendataphilly.org). As illustrated in Fig. \ref{fig2}, nDSMs are derived from Lidar point clouds. Firstly, noises in the point clouds are removed. Then the height values are rasterized into DSMs (digital surface models) with all the points, and DTMs (digital terrain models) with only the ground points. Finally, subtracting DTM from DSM gives nDSM. The classified point clouds are also ingredients for semantic maps when land cover maps are not available from open data sources. This is simply done by rasterizing the class labels. All the processed data are aligned and cropped into patches to make the final dataset.
	
	\subsection{Statistics of the GAMUS dataset}
	The introduced GAMUS dataset contains 11,507 tiles collected from five different cities: Oklahoma, Washington, D.C., Philadelphia, Jacksonville, and New York City. These image tiles are collected using the aforementioned data collection process, as shown in Fig. \ref{fig2}. Each RGB image tile has a corresponding nDSM map with a spatial size of 1024$\times$1024. We split all the image tiles into three subsets: the training set with 6,304 tiles, the validation set with 1,059 tiles, and the test set with 4,144 tiles. All the image pixels are annotated with six different land cover types, including 1. ground; 2. low-vegetation; 3. building; 4. water; 5. road; 6. tree. The height statistics provided in the nDSM are displayed in Fig. \ref{Hstat}. It can be seen from this figure, there is an obvious long-tailed distribution of the heights, i.e., the number of pixels with lower height values is significantly more than those with higher height values. We also show the statistics of the spatial distribution by averaging the height values at each image pixel across the whole dataset. We can see that the spatial distributions are different for the train, validation, and test subsets. 

    Furthermore, we also display the height distributions of different cities in Fig. \ref{HCity}. For all the cities, the height values obey clear long-tailed distributions. We also visualize the spatial patterns of the height maps by averaging all the height values for each position. We can see that the spatial patterns of different cities are quite different. In Fig. \ref{HCLS}, we present the distributions of semantic labels for different cities. This analysis reveals significant differences in the distribution of semantic objects across different cities.
	
	\begin{figure*}
		\centering
		\includegraphics[width=0.96\linewidth]{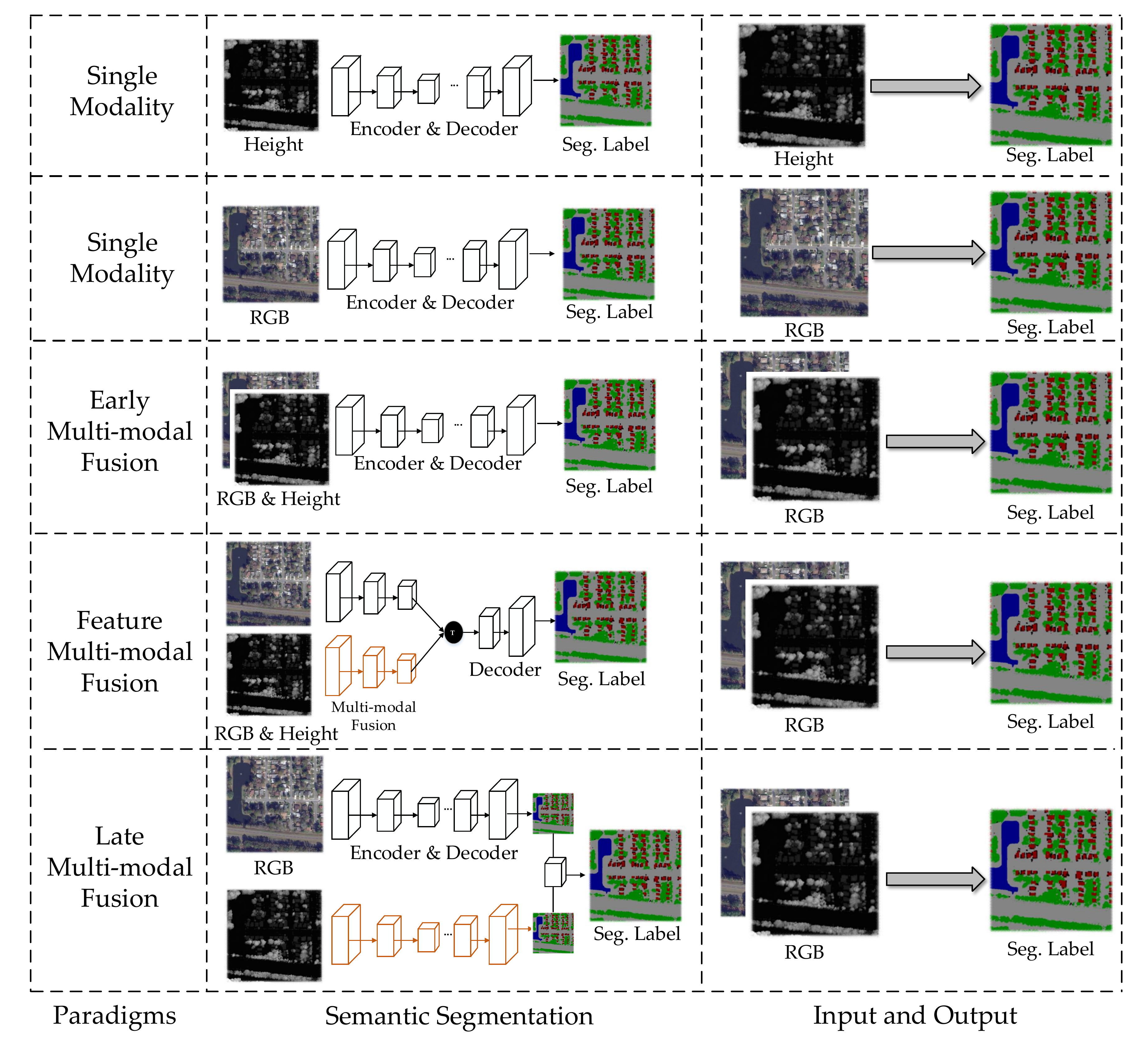}
		\caption{Illustration of five different multi-modal fusion paradigms, including 1) single modality, where only the RGB modality is used; 2) 1) single modality, where only the height modality is used; 3) multi-modal early-fusion, where image-level fusion is conducted; 4) multi-modal feature fusion, where features of different modalities are fused; 5) multi-modal late-fusion, where segmentation results from different modalities are combined.}
		\label{fig4}
	\end{figure*}
	
	\section{Methods}
    \label{Methods}
    Existing deep learning models for multi-modal semantic segmentation can be divided into CNN-based and Transformer-based networks. For CNN-based methods, the fusion in different layers has a great influence on the performance. For Transformer-based methods, the self-attention mechanism can be used to fuse multi-modal features at a token level \citep{xu2022multimodal}, which can be more effective than CNN-based methods. 
    \subsection{CNN-based Fusion Methods}
	For CNN-based fusion methods, we evaluate the performance using five different training paradigms as shown in Fig. \ref{fig4}. To formally define these training paradigms, we first introduce the notations of input modalities and networks. We use $\bm{X}_{rgb} \in \mathbb{R}^{3\times H\times W}$ and $\bm{X}_h \in \mathbb{R}^{H\times W}$ to denote the RGB input and height from nDSM, respectively. For the sake of simplicity, the encoder and decoder sub-networks for RGB and height modality are denoted by $f_{\mathbf{W}_{rgb}}^{e}$, $f_{\mathbf{W}_{rgb}}^{d}$ and $f_{\mathbf{W}_{h}}^{e}$, $f_{\mathbf{W}_{h}}^{d}$. With these notations, the five training paradigms can be formally expressed as follows.
	
	\begin{enumerate}
		\item \textbf{Single Modality (RGB)}. \\ $y_{s}^*=f_{\mathbf{W}_{rgb}}^{d}(f_{\mathbf{W}_{rgb}}^{e}(X_{rgb}))$.
        \item \textbf{Single Modality (Height)}. \\ $y_{s}^*=f_{\mathbf{W}_{h}}^{d}(f_{\mathbf{W}_{h}}^{e}(X_{h}))$.
		\item \textbf{Early Multi-Modal Fusion}. \\ $y_{s}^*=f_{\mathbf{W}_{}}^{d}(f_{\mathbf{W}_{}}^{e}(X_{rgb}\parallel X_h))$.
		\item \textbf{Feature-level Multi-Modal Fusion}. \\ $y_{s}^*=f_{\mathbf{W}_{}}^{d}(f_{\mathbf{W}_{rgb}}^{e}(X_{rgb}) \parallel f_{\mathbf{W}_{h}}^{e}(X_{h}))$.
		\item \textbf{Late Multi-Modal Fusion}.\\ $y_{s}^*=f_{\mathbf{W}_{rgb}}^{d}(f_{\mathbf{W}_{rgb}}^{e}(X_{rgb}))\, +f_{\mathbf{W}_{h}}^{d}(f_{\mathbf{W}_{h}}^{e}(X_{h}))$.
	\end{enumerate}
	
	Since the height information in the nDSM data is highly similar to the geometric information in the depth images, we select three RGB-D multi-modal learning methods for benchmarking. ShapeConv \citep{ShapeConv}, VCD \citep{xiong2020variational} aims at designing a cross-modality guided encoder to better fuse RGB features and depth information. FuseNet \citep{Hazirbas2016FuseNet} focuses on designing better multi-scale feature fusion architectures. 
 \begin{figure*}
		\centering
		\includegraphics[width=1\linewidth]{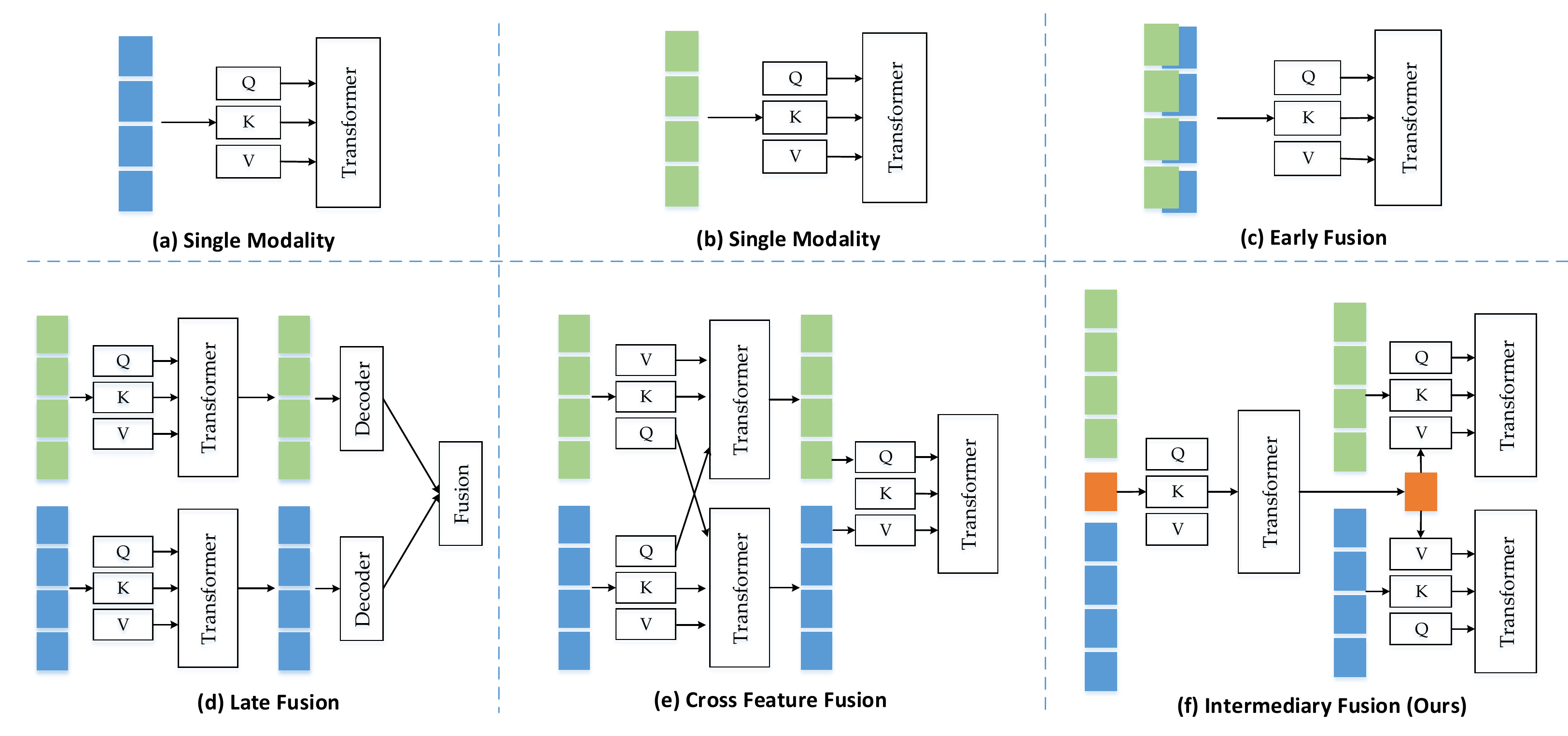} 
		\caption{Illustration of six different multi-modal fusion paradigms, including 1) single modality, where only the RGB modality is used; 2) single modality, where only the height modality is used; 3) multi-modal early-fusion, where image-level fusion is conducted; 4) multi-modal cross feature fusion, where features of different modalities are fused via cross-attention mechanism; 5) multi-modal late-fusion, where segmentation results from different modalities are combined; 6) Intermediary fusion, where the proposed TIMF module is illustrated.}
		\label{tfusion}
	\end{figure*}
	Considering the diversity of different multi-modal architecture designs, we also choose two RGB-T segmentation methods for performance evaluation on our GAMUS dataset. Multi-spectral fusion networks (MFNet) \citep{MFNet} takes RGB images and IR images as input and fuse them in the feature level for multi-modal learning. RTFNet \citep{RTFNet} uses two separate encoders for RGB and thermal modalities and fuses their intermediate features progressively to learn multi-modal representations. 
	
	For the performance evaluation of multiple modalities, we adapt three methods (MFNet, RTFNet, and FuseNet) to five types of multi-modal fusion paradigms to compare the performance. Specifically, for single modality settings, we modify them by removing one input modality and only remain a single RGB or height modality. For the early-fusion method, we change the model by simply stacking the RGB and height modality together as the multi-modal input. As for the feature-level fusion, we use the same network architectures as described in the original papers of the eight compared methods. For the late fusion setting, we only fuse the predicted segmentation results of each modality at the last layer of the segmentation head. Toward fair comparisons, we exploit the exact same network architectures and training hyper-parameters for methods trained in different experimental settings.
	
	\subsection{Transformer-based Fusion Methods}
    CMX \citep{liu2022cmx} introduces a cross-modal feature rectification module to calibrate the feature of one modality from another in both spatial- and channel-wise dimensions. SegFormer \citep{xie2021segformer} is used as its baseline segmentation architecture. Considering its effectiveness, we take CMX as our baseline method and adapt it to five different fusion strategies for performance comparison. Fig. \ref{tfusion} presents six types of fusion methods. The first five fusion methods including single modality, early fusion, late fusion, and cross-modal feature fusion \citep{liu2022cmx} are similar to the five CNN-based fusion strategies.
    Different from these five fusion methods, in this work, we propose the intermediary feature fusion module (illustrated in Fig. \ref{tfusion} (f)), which is described in the following section.

    To formally describe these Transformer-based fusion models, we first give a brief introduction to the widely-used vision Transformer (ViT). Given the input data $\bm{x}$, a Transformer encoder first projects the image patches into patch embeddings (denoted as $p(\cdot)$) and add position embeddings $\bm{x}_{pos}$ to enhance the position information. Then the input will further go through alternating layers of multi-head self-attention (MSA) and multi-layer perception (MLP) blocks. Layer Normalization (LN) and residual connection are applied before and after every block. The computation process of a Transformer block $\text{TL}(\bm{x})$ can be expressed as
    \begin{equation}
        \begin{split}
            \bm{x}_0 &= p(\bm{x}) + \bm{x}_{pos}, \\
            \bm{z}_k' &= \bm{z}_{k-1} + \text{MSA}(\text{LN}(\bm{x}_{k-1})), \\
            \bm{z}_k &= \bm{z}_k' + \text{MLP}(\text{LN}(\bm{z}_k')),
        \end{split}
    \end{equation}
    where $\bm{z}_k$ is the output features, and $k$ is the index of blocks.

    Let ${F}_{m}^e(\cdot)$ and $F_{m}^d(\cdot)$ be the encoder and decoder of the Segformer \citep{} networks, where $m$ could be either the RGB or the height modality. The single modality model with RGB input can be defined as $y_{s}^*=F^{d}(F^{e}(X_{rgb}))$. The computation for the height modality can be defined by simply replacing $X_{rgb}$ with $X_d$. The early fusion strategy can be expressed as $y_{s}^*=F^{d}(F^{e}(X_{rgb} \parallel X_d))$. Similarly, the late fusion model is formulated as
    \begin{equation}
     y_{s}^*=F_{rgb}^{d}(F_{rgb}^{e}(X_{rgb})) + F_h^{d}(F_d^{e}(X_{h})).
    \end{equation}
    The Transformer-based cross feature fusion is conducted by exchanging queries of different modalities in the MSA blocks. The other computation steps are the same as a Transformer block. For the sake of simplicity, we omit them and only present the computation of MSA, which can be expressed as:
    \begin{equation}
        \begin{split}
            &\bm{z}_{rgb} \longleftarrow \text{MSA}(\text{LN}(\underbrace{\bm{X}_{h},\bm{X}_{rgb},\bm{X}_{rgb}}_{Q,\quad K,\quad V})),\\
            &\bm{z}_h \longleftarrow \text{MSA}(\text{LN}(\underbrace{\bm{X}_{rgb},\bm{X}_{h},\bm{X}_{h}}_{Q,\quad K,\quad V})).
        \end{split}
    \end{equation}

    By this means, the multi-modal features are fused via cross-modal self-attention modules, which is more flexible than layer-wise feature fusion. 

    \subsection{Intermediary Multi-modal Fusion}
    Different from existing fusion strategies, in this work, we propose a novel Transformer-based intermediary multi-modal fusion method, i.e., the TIMF module. TIMF exploits an additional intermediary token $\bm{M}$ to adaptively combine RGB and height features at a token level via the self-attention mechanism. To be more specific, TIMF is designed in a hierarchical manner. As illustrated in Fig. \ref{tfusion} (f), the intermediary token first extracts global multi-modal features by feeding the intermediary token into an MSA module, which is similar to the $\text{CLS}$ token in ViT. Next, the output intermediary token $\bm{M}'$ is concatenated to the tokens of each modality respectively. Then, single-block ($k=1$) Transformer layers are used to further fuse the intermediary token $\bm{M}'$ with modality-specific tokens via self-attention modules. Formally, the first stage of TIMF takes $\bm{M}$ and tokens of both modalities as input to a single-block Transformer layer, which can be formulated as
    \begin{equation}
        \begin{split}
        \bm{z}_a &= \text{TL}(p(\bm{X}_{rgb}) \parallel p(\bm{X}_d) \parallel M),\\
        \bm{z}_a &= (\bm{z}_{rgb}) \parallel (\bm{z}_h) \parallel M'),
        \end{split}
    \end{equation}
    where $\bm{z}_a$ is the output feature. $\parallel$ is the concatenation operation. By slicing $\bm{z}_a$ according to the number of tokens, we can obtain the output features of RGB modality $\bm{z}_{rgb})$, height modality $\bm{z}_{h})$, and the output intermediary token $\bm{M}'$. Then, the final output features of the TIMF module are obtained by fusing $\bm{M}'$ with these two modalities individually using two Transformer layers. This can be formulated as 
    \begin{equation}
        \begin{split}
        \bm{z}_{rgb}' &= \text{TL}_{rgb}(p(\bm{z}_{rgb}) \parallel \bm{M}'),\\
        \bm{z}_{rgb}' &= (\bm{z}_{rgb}) \parallel \bm{M}'_{rgb}), \\
        \bm{z}_{h}' &= \text{TL}_{h}(p(\bm{z}_{h}) \parallel \bm{M}'),\\
        \bm{z}_{h}' &= ((\bm{z}_h) \parallel \bm{M}'_{h}).
        \end{split}
    \end{equation}
    
In practice, TIMF module can be used in a plug-and-play manner. We simply replace the TIMF module with the feature fusion module in CMX \citep{liu2022cmx}.

	\section{Experiments}
 \label{Experiments}
	The constructed dataset GAMUS in this paper can be publicly accessed at \url{https://github.com/EarthNets/Dataset4EO}. In this dataset, we have explicitly provided the official splits for training, validation, and test subsets. Providing a well-defined data set and the official split is crucial for reproducibility.

\subsection{Implementation details}
	\label{sec:setup}
	All the evaluated models are implemented using Pytorch and run on eight GeForce RTX 3090 GPUs. We use the public codebase provided by the original paper. For all the supervised multi-modal segmentation methods, 100 epochs are used for training. The input images are resized to 512$\times$512.
	The batch sizes are set to 16, and we use the same learning rate and optimizer as described in the original paper. For data augmentation, random flip and random crop are used for all the methods towards a fair comparison. On the GAMUS dataset, 50 epochs are used for the model training. The implemented dataset loading code can be found in \url{https://github.com/EarthNets/Dataset4EO}. For all the experiments, the mean Intersection over Union (mIoU) and IoU of each class are used as the evaluation metrics.

{To ensure reproducibility, we choose to make all the used source codes publicly available at \url{https://github.com/EarthNets/RSI-MMSegmentation}.
\begin{table*}[h!]
		\caption{Comparison results (Acc) of different multi-modal fusion methods on the GAMUS dataset for supervised semantic segmentation. }
		\label{CNN-ACC}
		\centering
  \scalebox{0.9}{
\begin{tabular}{c|c|c|c|c|c|c|c|c|c}
\hline
Paradigm                          & Methods                               & Modality & Ground & Vegetation & Building & Water  & Road   & Tree   & mAcc   \\ \hline \hline
\multirow{3}{*}{\makecell[c]{Single\\Modality}} & MFNet \citep{MFNet}                   & RGB      & 0.7030 & 0.4251     & 0.7696   & 0.3510 & 0.4242 & 0.7947 & 0.5779 \\
                                  & RTFNet \citep{RTFNet}                 & RGB      & 0.7370 & 0.5980     & 0.8873   & 0.2144 & 0.6236 & 0.8666 & 0.6545 \\
                                  & FuseNet \citep{Hazirbas2016FuseNet}   & RGB      & 0.3753 & 0.5104     & 0.8724   & 0.1045 & 0.6375 & 0.7887 & 0.5481 \\ \hline
\multirow{3}{*}{\makecell[c]{Single\\Modality}} & MFNet \citep{MFNet}                 & Height   & 0.6881 & 0.5304     & 0.7601   & 0.4078 & 0.4911 & 0.7502 & 0.6046 \\
                                  & RTFNet \citep{RTFNet}               & Height   & 0.8223 & 0.5537     & 0.8708   & 0.2513 & 0.6764 & 0.8186 & 0.6655 \\
                                  & FuseNet \citep{Hazirbas2016FuseNet} & Height   & 0.7821 & 0.4208     & 0.8173   & 0.6912 & 0.7455 & 0.4715 & 0.6547 \\ \hline
\multirow{4}{*}{\makecell[c]{Early\\Fusion}}    & MFNet \citep{MFNet}                   & RGBH     & 0.6968 & 0.5749     & 0.7875   & 0.4673 & 0.4917 & 0.8005 & 0.6365 \\
                                  & RTFNet \citep{RTFNet}                 & RGBH     & 0.7955 & 0.6521     & 0.8706   & 0.4695 & 0.6516 & 0.8135 & 0.7088 \\
                                  & FuseNet \citep{Hazirbas2016FuseNet}   & RGBH     & 0.8084 & 0.4178     & 0.9049   & 0.6758 & 0.8179 & 0.6807 & 0.7176 \\ \hline
                                  
\multirow{4}{*}{\makecell[c]{Feature\\Fusion}}  & MFNet \citep{MFNet}                   & RGBH    & 0.7211 & 0.5945     & 0.8020   & 0.5041 & 0.5298 & 0.8530 & 0.6674 \\
                                  & RTFNet \citep{RTFNet}                 & RGBH    & 0.7190 & 0.6732     & 0.8774   & 0.5975 & 0.7057 & 0.8533 & 0.7377 \\
                                  & FuseNet \citep{Hazirbas2016FuseNet}   & RGBH    & 0.4716 & 0.7558     & 0.9109   & 0.3184 & 0.6221 & 0.9444 & 0.6705 \\ \hline
                                  
\multirow{3}{*}{\makecell[c]{Late\\Fusion}}     & MFNet \citep{MFNet}                   & RGBH     & 0.7491 & 0.6037     & 0.8517   & 0.5530 & 0.5774 & 0.7831 & 0.6863 \\
                                  & RTFNet \citep{RTFNet}                 & RGBH     & 0.7798 & 0.6195     & 0.8941   & 0.5141 & 0.6953 & 0.8087 & 0.7186 \\
                                  & FuseNet \citep{Hazirbas2016FuseNet}   & RGBH     & 0.7239 & 0.3185     & 0.9526   & 0.0758 & 0.8571 & 0.8194 & 0.6245 \\ \hline
\end{tabular}}
	\end{table*}

 \begin{table*}[h!]
		\caption{Comparison results (IoU) of different multi-modal fusion methods on the GAMUS dataset for supervised semantic segmentation. }
		\label{CNN-IOU}
		\centering
  \scalebox{0.9}{
        \begin{tabular}{c|c|c|c|c|c|c|c|c|c}
        \hline
        Paradigm                          & Methods                             & Modality & Ground & Vegetation & Building & Water  & Road   & Tree   & mIoU   \\ \hline \hline
        \multirow{3}{*}{\makecell[c]{Single\\Modality}} & MFNet \citep{MFNet}                 & RGB      & 0.5560 & 0.3776     & 0.6776   & 0.2713 & 0.3036 & 0.6231 & 0.4682 \\
                                          & RTFNet \citep{RTFNet}               & RGB      & 0.6337 & 0.4326     & 0.7474   & 0.2197 & 0.5286 & 0.7026 & 0.5441 \\
                                          & FuseNet \citep{Hazirbas2016FuseNet} & RGB      & 0.5667 & 0.3439     & 0.5652   & 0.3958 & 0.4298 & 0.4326 & 0.4557 \\ \hline
        \multirow{3}{*}{\makecell[c]{Single\\Modality}} & MFNet \citep{MFNet}                 & Height   & 0.5199 & 0.3393     & 0.7382   & 0.1918 & 0.2479 & 0.7064 & 0.4572 \\
                                          & RTFNet \citep{RTFNet}               & Height   &0.5887	&0.3912	&0.7522	&0.1807	&0.4752	&0.7206	&0.5181 \\
                                          & FuseNet \citep{Hazirbas2016FuseNet} & Height   & 0.3753 & 0.5104     & 0.8724   & 0.1045 & 0.6375 & 0.7887 & 0.5481 \\ \hline
        \multirow{4}{*}{\makecell[c]{Early\\Fusion}}    & MFNet \citep{MFNet}                 & RGBH     & 0.5773 & 0.3990     & 0.7345   & 0.2511 & 0.3073 & 0.7056 & 0.4958 \\
                                          & RTFNet \citep{RTFNet}               & RGBH     & 0.6218 & 0.4502     & 0.7457   & 0.3771 & 0.5240 & 0.7083 & 0.5712 \\
                                          & FuseNet \citep{Hazirbas2016FuseNet} & RGBH     & 0.6359 & 0.3702     & 0.6914   & 0.5321 & 0.4336 & 0.6289 & 0.5487 \\
                                          \hline
        \multirow{4}{*}{\makecell[c]{Feature\\Fusion}}  & MFNet \citep{MFNet}                 & RGBH    & 0.6034 & 0.4480     & 0.7697   & 0.2563 & 0.3347 & 0.7517 & 0.5273 \\
                                          & RTFNet \citep{RTFNet}               & RGBH    & 0.6010 & 0.4431     & 0.7592   & 0.4507 & 0.5190 & 0.7226 & 0.5826 \\
                                          & FuseNet \citep{Hazirbas2016FuseNet} & RGBH    & 0.4318 & 0.4186     & 0.7514   & 0.2979 & 0.3937 & 0.6773 & 0.4951 \\
                                          \hline
        \multirow{3}{*}{\makecell[c]{Late\\Fusion}}     & MFNet \citep{MFNet}                 & RGBH     & 0.6278 & 0.4606     & 0.8031   & 0.2997 & 0.4013 & 0.7831 & 0.5626 \\
                                          & RTFNet \citep{RTFNet}               & RGBH     & 0.6209 & 0.4384     & 0.7576   & 0.4537 & 0.5494 & 0.7162 & 0.5894 \\
                                          & FuseNet \citep{Hazirbas2016FuseNet} & RGBH     & 0.5624 & 0.3042     & 0.7699   & 0.0748 & 0.3975 & 0.7396 & 0.4747 \\ \hline
        \end{tabular}}
	\end{table*}

\subsection{Multi-modal Learning Analysis}
{
To analyze the benefits of using multiple modalities for semantic segmentation, we compare the experimental results of using different data modalities as input in Table \ref{CNN-ACC}, Table \ref{CNN-IOU}, and Table \ref{TF}. Specifically, for CNN-based fusion methods, we present the accuracy and mIoU results in Table \ref{CNN-ACC} and Table \ref{CNN-IOU}. As for the Transformer-based methods, we provide the mIoU results of six different types of modality inputs in Table \ref{TF}. In the following, we analyze and discuss these extensive results from five different aspects.

\paragraph{1. Benefits of Multi-modal Learning} Despite the fact that early-fusion is the simplest multi-modal learning method, stacking RGB and height map as a four-channel input, can still obtain clearly better results than using a single RGB modality. As presented in Table \ref{CNN-IOU}, for CNN-based methods, using early-fusion, MFNet can improve the mIoU of RGB modality from 46.8\% to 49.6\%. FuseNet can obtain a 9\% improvement than only using the RGB data. These results reveal the effectiveness of multi-modal learning for the segmentation of RS images.
\begin{figure*}
		\centering
		\includegraphics[width=0.97\linewidth]{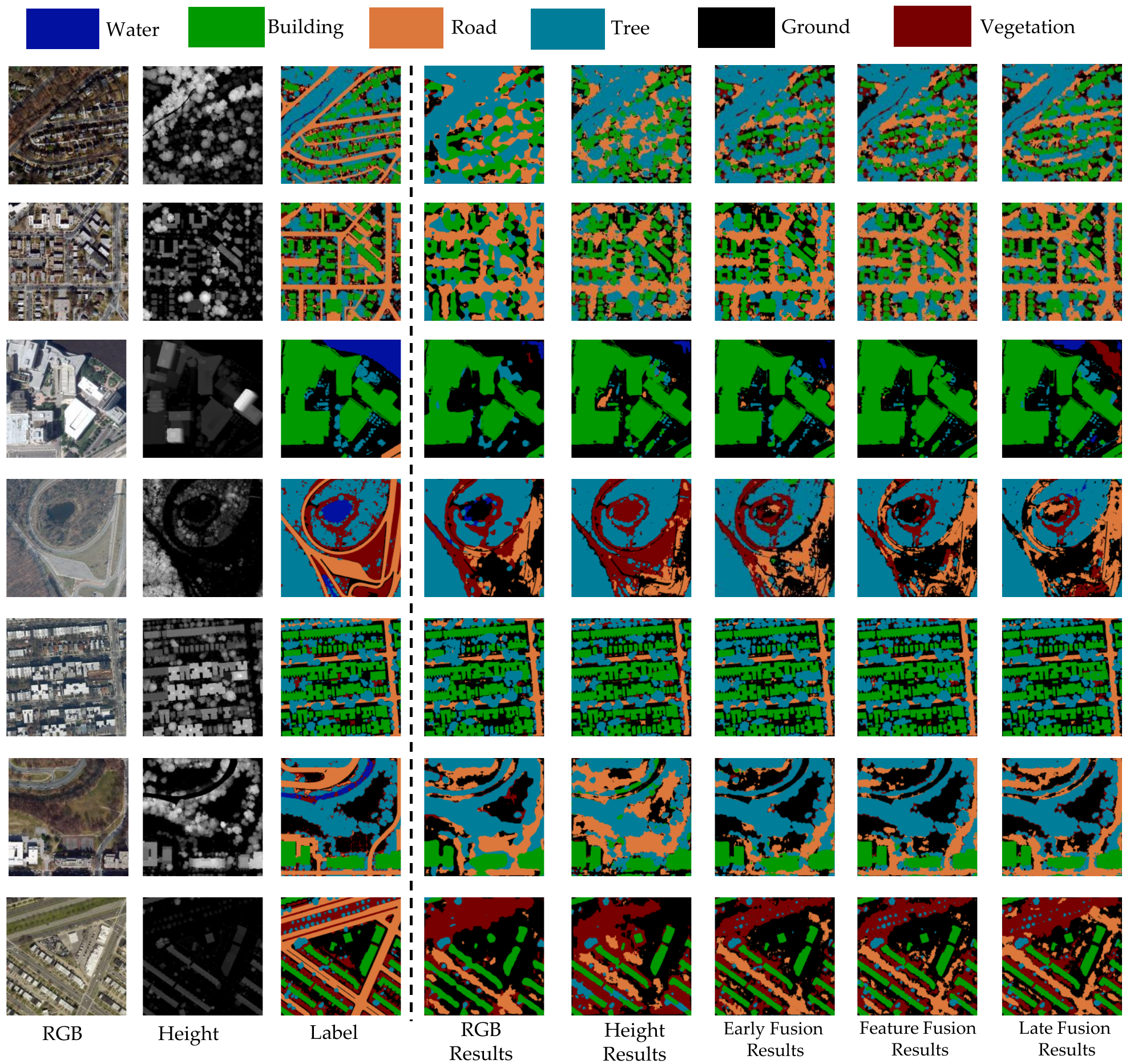}
		\caption{some qualitative visualization examples of segmentation results on the GAMUS dataset. The segmentation results of using different multi-modal fusion strategies are visualized for a clear comparison.}
		\label{modala}
\end{figure*}
Furthermore, by exploiting more sophisticated feature-level multi-modal fusion methods, the performance can be further improved. For example, MFNet with a feature-fusion strategy can obtain a 6\% improvement to using a single RGB modality. However, surprisingly, for FuseNet, the performances of the simple early-fusion methods may outperform feature-level fusion methods. In general, from the results, we can clearly see that using both the RGB and height modalities can obtain much better results compared to using single modalities. 

Comparing the results of using RGB and RGBH data, tree and building are the semantic classes with the greatest performance boost when the extra height modality is used. This makes sense because trees and buildings are objects with clearly higher height values. Comparing the results of using height and RGBH data, there is significant performance improvement. Specifically, ground and vegetation are the semantic classes with the greatest performance boost when the extra RGB modality is used. As ground and vegetation always have the lowest height data, they are not distinguishable if RGB texture is not used. The results also support that fusing multiple modalities is the key to improving the segmentation performance of remote sensing data. 

For Transformer-based multi-modal fusion methods, similar conclusions can be derived. As presented in Table \ref{TF}, a simple early-fusion method can boost the mIoU of using the single RGB data from 69\% to 75\%. Using the proposed TIMF module can achieve a 10\% performance improvement, which is significant for segmentation tasks. 
When looking into the results of different semantic classes, it can be found that the performance of five of the classes can be improved obviously. 

\paragraph{2. Comparison of Different Fusion Strategies} 
When comparing the performance of using the RGB data and height data, we can find that in most cases, using RGB performs better than using the height modality for the pixel-wise segmentation task. The reason is that RGB can provide rich textures of different semantic objects, which are important for learning discriminative representations. Nevertheless, using the height modality can clearly obtain better performance on the \textbf{building} and \textbf{tree} classes, as they are more sensitive to the geometry information. Thus, by designing effective multi-modal learning methods, complementary features can be learned and the performance can be largely improved. 

For CNN-based methods, among different multi-modal fusion strategies, we find that feature-fusion and late-fusion are generally more effective than early-fusion.
This makes sense because more diverse representations can be learned from complementary modalities with more sophisticated fusion methods. By comparing the results in Table \ref{CNN-IOU}, we can see that, the performance of late-fusion is relatively higher than early-fusion and feature-fusion methods in general. Some qualitative results are visualized in Fig. \ref{modala}. The segmentation results of using different multi-modal fusion strategies are visualized for a clear comparison. It can be seen that feature-fusion and late-fusion models can obtain greatly better segmentation results.

As for Transformer-based methods, it can be seen from the results in Table \ref{TF} that late-fusion can even outperform the cross-attention-based fusion method. This shares a similar conclusion with CNN-based methods. Late-fusion works surprisingly well for the RGBH segmentation task. These insights can be helpful for future research to design more effective multi-modal learning models. For Transformer-based fusion methods, we visualize the segmentation maps of different multi-modal fusion strategies in Fig. \ref{modalb}. 

\begin{table*}[h!]
		\caption{Comparison results (IoU) of different fusion strategies on the GAMUS dataset for Transformer-based semantic segmentation methods. }
		\label{TF}
		\centering
  \scalebox{0.9}{
			\begin{tabular}{c|c|c|c|c|c|c|c|c|c}
				\hline \hline
				Paradigm & Methods & \multicolumn{1}{l|}{Modality} & Ground & Vegetation & Building & Water & Road  & \multicolumn{1}{l|}{Tree} & mIoU  \\ \hline
				{\makecell[c]{Single\\Modality}}
				&CMX \citep{liu2022cmx}      & RGB  & 0.7265 & 0.6047     & 0.7653   & \textbf{0.6988} & 0.6859 & 0.6790 & 0.6934 \\ \hline
				
				{\makecell[c] {Single\\Modality}}
				&{CMX} \citep{liu2022cmx}      & Height  & 0.6672 & 0.4816     & 0.8271   & 0.4203 & 0.5893 & 0.8024 & 0.6313 \\ \hline
				    
				{\makecell[c]{Early\\Fusion}}
				&CMX \citep{liu2022cmx}      & RGBH & 0.7928 & 0.6654     & 0.8236   & 0.6876 & 0.7273 & 0.8001 & 0.7495 \\ \hline

                {\makecell[c]{Late\\Fusion}}
				&CMX \citep{liu2022cmx}      & RGBH    & 0.7976 & 0.6763     & 0.8438   & 0.6827 & 0.7281 & 0.8249 & 0.7589 \\ \hline

                {\makecell[c]{Cross Feature\\Fusion}}
				&CMX \citep{liu2022cmx}      & RGBH    & 0.7827 & 0.6710     & \textbf{0.8453}   & 0.6770 & 0.7127 & \textbf{0.8250} & 0.7523 \\ \hline

                {\makecell[c]{Intermediary\\Fusion}}
				&TIMF (Ours)      & RGBH    & \textbf{0.8023} & \textbf{0.6797}     & 0.8452   & {0.6955} & \textbf{0.7368} & 0.8232 & \textbf{0.7638} \\ \hline
				
				\hline \hline
			\end{tabular}}
	\end{table*}
 \begin{table*}[h!]
		\caption{Comparison results (IoU) of different fusion strategies on the GAMUS dataset for supervised semantic segmentation. }
		\label{SOTA}
		\centering
        \scalebox{0.92}{
			\begin{tabular}{c|c|c|c|c|c|c|c|c}
				\hline \hline
				Methods & \multicolumn{1}{l|}{Modality} & Ground & Vegetation & Building & Water & Road  & \multicolumn{1}{l|}{Tree} & mIoU  \\ \hline

				MFNet \citep{MFNet}                 & RGBH    & 0.6034 & 0.4480     & 0.7697   & 0.2563 & 0.3347 & 0.7517 & 0.5273 \\ \hline
				
				RTFNet \citep{RTFNet}               & RGBH    & 0.6010 & 0.4431     & 0.7592   & 0.4507 & 0.5190 & 0.7226 & 0.5826 \\ \hline

				FuseNet \citep{Hazirbas2016FuseNet} & RGBH    & 0.4318 & 0.4186     & 0.7514   & 0.2979 & 0.3937 & 0.6773 & 0.4951 \\ \hline
                
				MFNet-ShapeConv \citep{ShapeConv}         & RGBH     & 0.6172 & 0.4655     & 0.7693   & 0.2190 & 0.5485 & 0.7320 & 0.5586 \\ \hline
                
				MFNet-VCD \citep{xiong2020variational}    & RGBH    & 0.6401 & 0.4560     & 0.7801   & 0.4595 & 0.5203 & 0.7261 & 0.5970 \\ \hline
                
                CMX \citep{liu2022cmx}      & RGBH    & 0.7827 & 0.6710     & \textbf{0.8453}   & 0.6770 & 0.7127 & \textbf{0.8250} & 0.7523 \\ \hline

                TIMF (Ours)      & RGBH    & \textbf{0.8023} & \textbf{0.6797}     & 0.8452   & \textbf{0.6955} & \textbf{0.7368} & 0.8232 & \textbf{0.7638} \\ \hline
				
				\hline \hline
			\end{tabular}}
	\end{table*}
\paragraph{3. Comparison between CNN and Transformer-based Methods.}
From the experimental results, it can be clearly seen that Transformer-based segmentation models can achieve much better performance than CNN-based methods. Even using a single RGB modality, the performance is much higher than CNN-based methods. We attribute the performance improvement to two reasons. One is that CMX \citep{liu2022cmx} using SegFormer \citep{xie2021segformer} as a baseline can achieve better performance owing to its global spatial context and self-attention mechanism for learning better representations. The second reason is that Transformer-based methods can fuse multi-modal features at a more flexible token level.

\paragraph{4. Comparisons of State-of-the-art Multi-modal Learning Models.} We compare seven existing methods that are designed for the multi-modal segmentation task. The results in Table \ref{SOTA} indicate that designing better feature-fusion methods is useful to improve the segmentation performance. The results from ShapeConv \citep{ShapeConv} and VCD \citep{xiong2020variational} reveal that making better use of the geometry information in the height modality can help improve the segmentation performance. Compared with existing methods, TIMF module achieves the best mIoU performance, which demonstrates the effectiveness of the proposed method.

The multi-modal semantic segmentation task is still under-explored, and the value of the extra height modality is still not fully utilized. More research efforts are required to further improve the performance.
}

    \begin{figure*}
		\centering
		\includegraphics[width=0.97\linewidth]{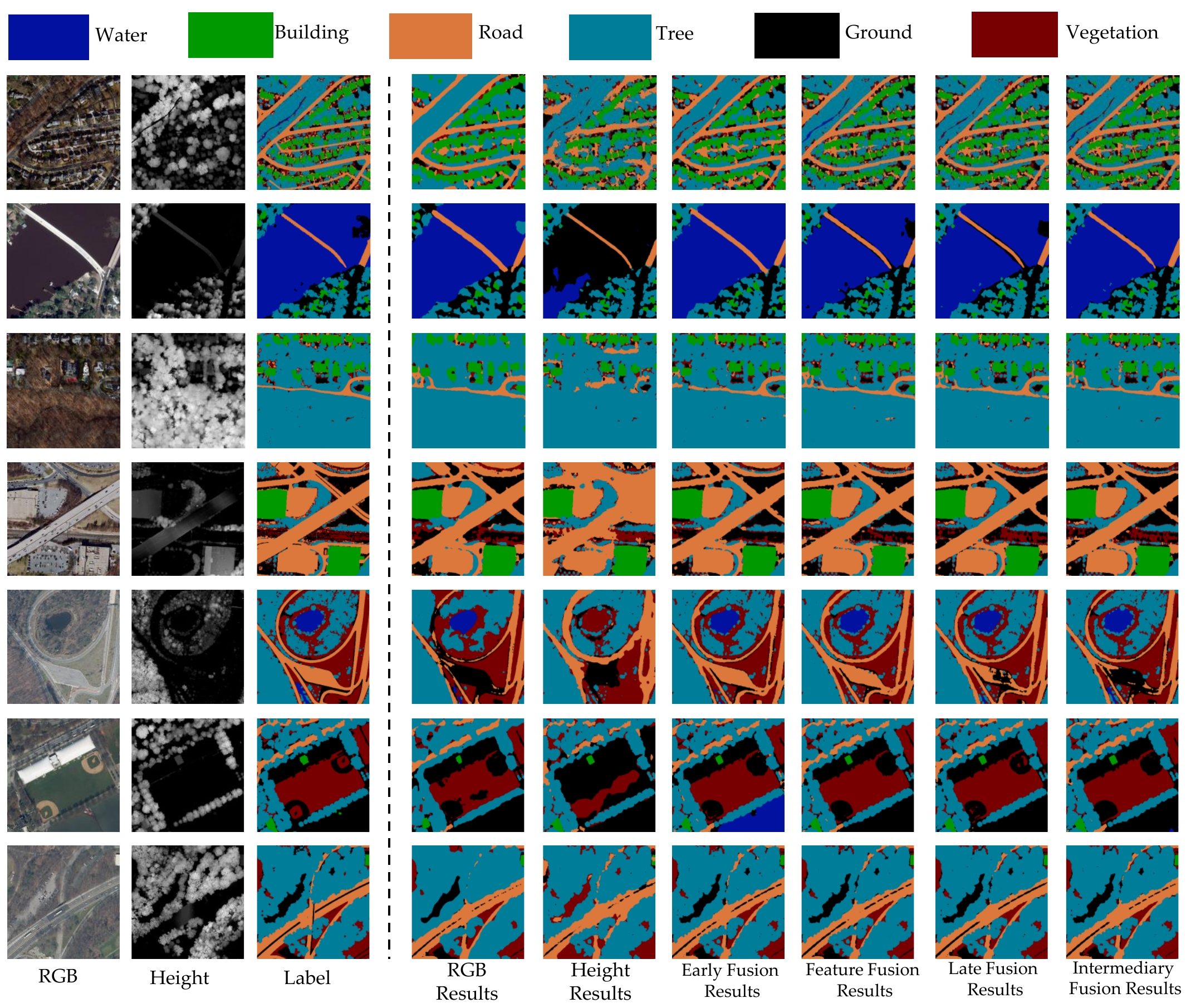}
		\caption{some qualitative visualization examples of segmentation results on the GAMUS dataset. The segmentation results of using different multi-modal fusion strategies are visualized for a clear comparison.}
		\label{modalb}
    \end{figure*} 
	
\section{Conclusion}
	\label{discuss}
    In this work, we focus on the multi-modal segmentation task, where two modalities (RGB and nDSM (height)) are jointly used to improve the segmentation performance. It is still an under-explored field in remote sensing due to the lack of large-scale datasets and unified benchmarks. This leads to difficulties in comparing the effectiveness of different algorithms. Thus, it is still not clear which type of fusion method is suitable for remote sensing data. To cope with these problems, in this work, we introduce a new remote-sensing benchmark dataset for multi-modal semantic segmentation based on RGB-Height (RGB-H) data. Towards a fair and comprehensive analysis of existing methods, the proposed benchmark consists of a large-scale dataset including co-registered RGB and nDSM pairs and pixel-wise semantic labels and a comprehensive evaluation and analysis of existing multi-modal fusion strategies. Both convolutional networks and Transformer-based networks are studied on the proposed dataset. Additionally, we propose a novel Transformer-based intermediary multi-modal fusion (TIMF) module to improve the semantic segmentation performance by adaptively fusing the RGB and Height modality. Extensive analyses of those methods are conducted and valuable insights are provided through the experimental results. We believe that GAMUS is an important step towards a fair and comprehensive benchmarking on multi-modal learning with RGB and geometric modality in remote sensing and earth observation.
		
\section*{Acknowledgement}
The work is jointly supported by German Federal Ministry for Economic Affairs and Climate Action in the framework of the "national center of excellence ML4Earth" (grant number: 50EE2201C), by the German Federal Ministry of Education and Research (BMBF) in the framework of the international future AI lab "AI4EO -- Artificial Intelligence for Earth Observation: Reasoning, Uncertainties, Ethics and Beyond" (grant number: 01DD20001) and by the Helmholtz Association through the Framework of the Helmholtz Excellent Professorship ``Data Science in Earth Observation - Big Data Fusion for Urban Research''(grant number: W2-W3-100).
\bibliographystyle{cas-model2-names}
\bibliography{refs}
\end{document}